\documentclass[11pt,a4paper]{article}
\usepackage[hyperref]{emnlp2018}
\usepackage{times}
\usepackage{latexsym}

\usepackage{bigfoot}

\usepackage{amsfonts}
\usepackage{amsmath}
\usepackage{graphicx}

\usepackage{balance}

\usepackage[utf8]{inputenc}
\usepackage[english]{babel}

\usepackage{url}

\aclfinalcopy % Uncomment this line for the final submission

\title{What is wrong with style transfer for texts?}

\author{Alexey Tikhonov \\
  Yandex  \\
  Karl-Liebknecht strasse 1, Berlin \\
  {\tt altsoph@gmail.com} \\\And
  Ivan P. Yamshchikov \\
  Max Planck Institute for Mathematics in the Sciences \\
  Inselstrasse 22, Leipzig \\
  {\tt ivan@yamshchikov.info} \\}

\date{}

\begin{document}
\maketitle
\begin{abstract}
A number of recent machine learning papers work with an automated style transfer for texts and, counter to intuition, demonstrate that there is no consensus formulation of this NLP task. Different researchers propose different algorithms, datasets and target metrics to address it. This short opinion paper aims to discuss possible formalization of this NLP task in anticipation of a further growing interest to it.
\end{abstract}

\section{Introduction}

Arguably, a growing interest to style transfer algorithms for texts is motivated by \citet{Gatys}, where style transfer was developed for images in a very convincing manner. \citet{Gatys} visualize the information at different processing stages in the convolutional neural network (CNN) by reconstructing the input image out of the network’s responses in a particular layer. This approach allowed the authors to avoid the burden of explicit style definition and conclude that in higher layers of the network, detailed pixel information is lost while the high-level content of the image is preserved. At the same time, discarding information on the global arrangement of the scene one can reconstruct the style of the input image from a style representation built on different subsets of CNN layers. This effective information decomposition, which makes high level image information explicit, seems to be an internal property of CNNs optimized for object recognition, but is not available for texts so far. In \citet{Zhang33} authors apply CNN to the task of text understanding and this promises a chance for a similar semantic-stylistic decomposition for texts in future, however at the moment it is difficult due to a number of reasons. First, textual information does not have the continuity that is characteristic for the images. Second, there is no 'characteristic' scale on which one can observe solely stylistic information: one can not say that the style of the text is determined on the level of letters but does not have anything to do with words or collocations. Finally, one usually speaks about the stylization of sentences rather than longer patterns of text. This inevitably implies a significantly lower amount of stylistic information available to the system in every separate piece of input. However, these difficulties as well as an absence of a clear consensus definition of the text style somehow do not hinder the intuitive understanding of the style transfer problem in the context of texts. This contradiction between an intuitive nature of the problem and its' formal complexity results in a number of research projects that not only look at the problem from very different angles but, in fact, even look on different problems giving them the same name. Further we list a number of different contributions that can be roughly split in three major groups that have very different understanding of this problem.

\section{Related work}

Let us try to systematize current approaches to the style transfer for texts. 

\subsection{Ad-hoc defined style classes}

Despite the fact that a sentiment of a sentence is not equivalent to its style there is a number of works that focus on the sentiment of the text specifically. In \citet{li}, for example, the authors estimate {\em the quality of the style transfer} with binary sentiment classifier trained on the same corpus of Yelp and Amazon reviews that is used for the training of the 'style'-transfer system. There are multiple results in this field, see \citet{Kabbara}, or \citet{Xu2}.  We suggest to call this understanding of the text style {\em ad-hoc}, since here the notion of the style is rigorously reverse-engineered out of a given training dataset.  \citet{Ficler} and especially \citet{Fu} generalize this ad-hoc approach in a clear and legitimate way. \citet{Fu} suggest to define style as a set of measurable categorial and/or continuos parameters. One trains a classifier for every parameter that comprises the style of the texts and then tests the resulting output with this pre-trained set of classifiers, using the percentage of correctly classified sentences as a measure of style transfer success. Subjectively, ad-hoc approaches tend to be extremely useful for a number of industrial tasks. This is partially due to a clear, applied problem set-up and partially to an exceeding number of human-supervised natural language datasets that could be used in the ad-hoc style-transfer setup. Enhanced with human peer-reviews as in \citet{TY} they can be very illustrative. However, since methods of this type do not imply any holistic and non-contradictory notion of style, talking about sentiment transfer or, say, text summarization instead of a 'style transfer' would make more sense in this ad-hoc paradigm.

\subsection{NMT approaches}

The idea behind these works is to define two different styles as two different languages and thus reduce the problem of style transfer to the problem of neural machine translation (NMT). A typical example of this approach could the so-called \textit{'style of the time'} (see \citet{Hughes}). \citet{xu} or \citet{Jhamtani} use parallel corpora of Shakespeare in original and modern language to train an automated 'shakespearizator'. \citet{Carlson} use parallel bible translations and discuss the results in the context of automated simplification that "can easily be viewed as style transfer". The use of such methods in practice is seriously hindered due to the deficit of parallel datasets with equivalent semantics and different stylistics. In \citet{Jang} authors address this problem trying to find an automated method for parallel corpora generation. In \citet{Rao} a dataset for formality style transfer is introduced alongside with the benchmarks and target metrics in the context of NMT. The problem of aligned dataset will stay characteristic for NMT approaches: \citet{Xu3} lists seven styles of language but immediately gives a disclaimer that 'There are certainly more than seven language styles as there are more than seven wonders in the world.' Each further 'wonder' would demand a separate aligned training corpus which makes NMT-approaches relatively labour-extensive. 

\subsection{Post-NMT approaches}

These are the approaches that follow the logic of NMT methods. The researchers agree that the notion of style should not be fragmentized as in ad-hoc approaches but should rather be extracted out of the corpora automatically. The same analogue of different styles being different languages holds here. However, understanding a limiting effect associated with the deficit of parallel language datasets, the researchers try to find work-arounds inspired by recent zero-shot NMT techniques, GANs, ect. In the last several years this post-NMT view started to get momentum, see \citet{Artetxe}, \citet{Han},  \citet{Shen}, \citet{Zhang2}, \citet{Prabhumoye}, \citet{Zhang1}. All of the methods try to obtain some latent representations that would correspond to stylistics and semantics separately (similarly to the information decomposition that we have for images). This might be done in a number of ways: 
\begin{itemize}
\item Obtaining regularized or somehow aligned embeddings for words or sentences and segmenting embedding state-space into the semantic and stylistic sections;
\item Using double transfer (there-and-back) as a target for the quality of the style transfer method;
\item Training a stylistic discriminator.
\end{itemize}

\section{Text style transfer}\label{formulation}

This brief overview of current approaches to style transfer definition arises a number of questions. The most interesting one is if there is a connection between the implied intuitive understanding of style and semantics. Indeed, ad-hoc defined style classes do not imply that the semantic part of the sentence should not be altered after a style-transfer. For example, if one assumes that sentiment is a stylistic feature, as some of the authors listed above do, that might lead to the following contradiction. Let us optimize a loss function that corresponds to an effective information decomposition of semantics and style on the dataset $X$ that looks like
\begin{equation}
\label{loss}
\mathcal{L}_{total} (X)= A_{semantic} (X) + A_{stylistic} (X),
\end{equation}
where $A$ stands for some measure of accuracy of style transfer procedure. 

Generally, one would expect to say that "this place has great candy" and "this place has awful candy" are two stylistically identical examples that have different semantics. However, if we treat sentiment as a stylistic feature, we have to assume that either $ A_{stylistic} (X)$ is a function of $A_{semantic} (X)$ (which means that it does not make sense to talk about style transfer anymore) or that there exists an effective information decomposition such that $A_{semantic} (X)$ and $A_{stylistic} (X)$ are independent functions of $X$. Indeed, in our example such decomposition could be obtained if one assumes that "this place has candy" is the part of the sentence that contains all semantically significant information whereas "great" is the part of the sentence that is responsible for its 'style'. However, the existence of such decomposition would mean that our final solution would inevitably be unstable, having equal resulting loss for sentences with different degree of semantics preservation and stylistic accuracy. Moreover, depending on how one measures $A_{semantic} (X)$ and $A_{stylistic} (X)$ there might be multiple reformulations with the same final loss $\mathcal{L}_{total} (X)$. For example "this place has average candy", "that fella sells awful caramels", "those restaurants serve horrific bonbons" might end up having comparable losses. Intuitively one would perceive "that fella sells awful caramels" and "those restaurants serve horrific bonbons" as two clear examples of semantically equivalent and stylistically different sentences, but this is not so under the assumption that sentiment is a stylistic attribute. One also has to mention that due to the language multimodality under such decomposition assumptions the sentence "this place has cotton candy" would be measured as semantically equivalent to the "this place has great candy" and stylistically could be assigned to the neutral sentiment. This, if we use the loss from Equation \ref{loss}, might make it preferable to the sentence "that fella sells awful caramels" (the sentiment part for the latter sentence might not compensate for it's semantic difference with "this place has great candy"). It would also make it equivalent to "this place has average candy", despite the common sense intuition that says that these two sentences are equivalent in style but are semantically different. This simple experiment illustrates a general weakness of ad-hoc approaches that have to define a loss under certain predefined information decomposition assumptions that are anything but trivial. This does not hinder applications of ad-hoc approaches to task-specific NLP problems but one has to keep this issue in mind and control for it. 

NMT approaches, on the other hand, do not have the loss-decomposition difficulty but raise some other questions. The most prominent questions here are:
\begin{itemize}
\item Can one say that any two semantically aligned corpora define a valid stylistic pair? Is, say, a style transfer problem from 'narcissistic' to 'academic' a legitimate style transfer task? 
\item Can there be an overlap of styles within one corpus? How can we control that?
\item Should every sentence in the corpus have a clear stylistic component? Can we remove or mix the sentences without a clear stylistic component in our corpus? Will such operation change the resulting style transfer algorithm?
\end{itemize}

All these questions are relevant for post-NMT methods as well, since the crucial difference between the two is the semantic alignment of two corpora, yet the style is latent content distribution across different text corpora both in NMT and post-NMT methods.

Keeping these ideas in mind we would suggest two clear criteria for a style-transfer task:
\begin{itemize}
\item Style is an integral component of text that allows it to be assigned to a certain category or a sub-corpus. The choice of such categories or sub-corpora is to be task-dependent.
\item Style has to be 'orthogonal' to semantics in the following way: any semantically relevant information could be expressed in any style.
\end{itemize}

Under these two criteria style transfer becomes a parallel shift with respect to a certain style coordinate, and latent representations are aligned to perform style transfer. Lexicon and sentence structure are two main tools for such shifts. Let sentence $X$ and its reformulation $\tilde{X}$ be represented as two points in state space $\mathbb{R}^n \bigotimes \mathbb{S}$, where $\mathbb{S}$ corresponds to the style dimension and $\mathbb{R}^n$ is a space of semantic embeddings. The loss for the problem of style transfer is defined as
\begin{equation}
\label{loss2}
\mathcal{L}_{total} (X, \tilde{X})= D_{\mathbb{R}^n}(X, \tilde{X}) + D_{S} (X, \tilde{X}),
\end{equation}
where $D_{\mathbb{R}^n}$ and $D_{S}$ are the notions of distances obtained our of a given corpus. Important part of this formalization is that we do not impose any structure over $S$. This structure is to be obtained through the exploration of a given corpus and the transferred sentences from one style match example sentences from the other style as a population.

\section{Conclusion}

In this short opinion paper we have listed different approaches to the problem of style transfer. We have broadly classified them into three main groups: ad-hoc approaches that use sets of pre-defined metrics for style, NMT approaches that use semantically aligned corpora and treat style as an integral property of a text that should not be formalized explicitly and post-NMT approach that are rooted in the same notion of non-explicit style formulation but try to avoid the formation of semantically aligned parallel corpora. We propose to distinguish ad-hoc tasks from NMT and post-NMT style-transfer tasks. We also suggest to talk about style transfer under the assumption of certain orthogonality between style and semantics that can be attained under NMT and post-NMT approaches and is unattainable in ad-hoc perspective.

\bibliography{shortemnlp2018}
\bibliographystyle{acl_natbib_nourl}

\end{document}